 \documentclass[preprint,10pt,3p]{elsarticle}

\usepackage{amssymb}
\usepackage{amsmath}

\usepackage{threeparttable}
\usepackage{multirow}
\usepackage{makecell}

\usepackage{eqparbox} 
\usepackage{array}

\journal{Ultrasonics}

\begin{document}

\begin{frontmatter}

\title{Switchable deep beamformer for high-quality and real-time passive acoustic mapping} 

\author[1]{Yi Zeng}
\author[2]{Jinwei Li}
\author[1]{Hui Zhu}
\author[4]{Shukuan Lu \corref{cor1}}
\ead{xjtusklu@mail.xjtu.edu.cn}
\author[2,3]{Jianfeng Li \corref{cor1}}
\ead{lijf1@shanghaitech.edu.cn}
\author[1,5,6]{Xiran Cai \corref{cor1}}
\ead{caixr@shanghaitech.edu.cn}

\affiliation[1]{organization={School of Information Science and Technology, ShanghaiTech University},
    city={Shanghai},
    postcode={201210}, 
    country={China}}

\affiliation[2]{organization={
Gene Editing Center, School of Life Science and Technology, ShanghaiTech University},
    city={Shanghai},
    postcode={201210}, 
    country={China}}

\affiliation[3]{organization={State Key Laboratory of Advanced Medical Materials and Devices, ShanghaiTech University},
    city={Shanghai},
    postcode={201210}, 
    country={China}}  
                
\affiliation[4]{organization={Key Laboratory of Biomedical Information Engineering of Ministry of Education, Department of Biomedical Engineering, School of Life Science and Technology, Xi’an Jiaotong University},
    city={Xi’an},
    postcode={710049}, 
    country={China}} 

\affiliation[5]{organization={Shanghai Engineering Research Center of Intelligent Vision and Imaging, ShanghaiTech University},
    city={Shanghai},
    postcode={201210}, 
    country={China}} 

\affiliation[6]{organization={Key Laboratory of Biomedical Imaging Science and System, Chinese Academy of Sciences},
    city={Shenzhen},
    postcode={518055}, 
    country={China}} 

\cortext[cor1]{Corresponding author}

\begin{abstract}
Passive acoustic mapping (PAM) is a promising tool for monitoring acoustic cavitation activities in the applications of ultrasound therapy. Data-adaptive beamformers for PAM have better image quality compared to the time exposure acoustics (TEA) algorithms. However, the computational cost of data-adaptive beamformers is considerably expensive. In this work, we develop a deep beamformer based on a generative adversarial network, which can switch between different transducer arrays and reconstruct high-quality PAM images directly from radio frequency ultrasound signals with low computational cost. The deep beamformer was trained on the dataset consisting of simulated and experimental cavitation signals of single and multiple microbubble clouds measured by different (linear and phased) arrays covering 1--15 MHz. We compared the performance of the deep beamformer to TEA and three different data-adaptive beamformers using the simulated and experimental test dataset. Compared with TEA, the deep beamformer reduced the energy spread area by 18.9\%--65.0\% and improved the image signal-to-noise ratio by 9.3--22.9 dB in average for the different arrays in our data. Compared to the data-adaptive beamformers, the deep beamformer reduced the computational cost by three orders of magnitude achieving 10.5 ms image reconstruction speed in our data, while the image quality was as good as that of the data-adaptive beamformers. These results demonstrated the potential of the deep beamformer for high-resolution monitoring of microbubble cavitation activities for ultrasound therapy.
\end{abstract}


\end{frontmatter}



\section{Introduction}
\label{sec:introduction}
Passive acoustic mapping (PAM) is a promising tool for localizing acoustic cavitation sources to monitor and guide ultrasound therapies. This imaging mode uses passively received acoustic emissions originated from a region of interest using an ultrasonic array, and allows continuous monitoring of cavitation activities without interrupting the therapeutic pulses~ \cite{MASIERO2022}. PAM has demonstrated its efficacy in monitoring acoustic cavitation activities during thermal ablation treatments \cite{Jensen2012,Jensen2013}, histotripsy \cite{bader2017} and microbubble (MB) mediated ultrasound treatments for drug delivery \cite{DiIanni2019}, and blood-brain barrier opening in both animal models and human subjects \cite{Bae2023ASA}.

Time exposure acoustics (TEA) \cite{Gyongy2010} based on the ‘delay and sum’ (DAS) operation and its extensions in the frequency domain \cite{Haworth2017}, are efficient for PAM. However, the low spatial resolution and artifacts stemming from the interference among multiple cavitation sources undermine the quality of the images reconstructed by these methods and the accuracy of localizing cavitation sources. To overcome these limitations, data-adaptive algorithms for PAM including robust Capon beamformer (RCB) \cite{Coviello2015}, Eigenspace-based robust Capon beamformer (EISRCB) \cite{Lu2018}, robust beamforming by linear programming (RLPB) \cite{Lyka2018} and dual apodization with cross-correlation combined with RCB (DAX-RCB) \cite{Lu2020} have been proposed. Using the data-adaptive algorithms, substantial artifacts suppression and improved spatial resolution for PAM have been demonstrated. However, these methods are highly computationally intensive, primarily because of the process of finding the weights for the sensors for each pixel. 
Recently, a data-adaptive spatial filtering approach is integrated with the PAM beamformer \cite{haworth2023} to reduce image artifacts. While the computational time is two orders of magnitude less than RCB, the threshold of the spatial filter depends on the total number of cavitation sources which is difficult to estimate accurately in experiments, because the dynamics of cavitation bubbles are changing over time. 
In addition, the minimum variance (MV) beamformer and the improved hybrid MV and delay-multiply-and-sum (MV-DMAS) has demonstrated their real-time capability, but a large part of the artifacts in deeper imaging depth still degrade the image quality \cite{Zhang2024a,Zhu2024}.
Therefore, it is still important to investigate new approaches comprising both fast image reconstruction speed and good image quality for PAM.

Deep learning based beamformers have been proposed in the ultrasound imaging field for various purposes, including fast and high quality imaging \cite{Luijten2019}, spatial correlation estimation \cite{Wiacek2020}, color flow imaging \cite{Nahas2020}, super-resolution imaging of micro-vessels \cite{Sloun2021} and passive cavitation imaging \cite{SharahiDeepPAM2023}.  While these previous works showed the success of neural networks in tackling multiple challenges in various ultrasound imaging modes, a common limitation in these related studies is that the neural networks trained may only apply for a specific transducer array. A recent work for suppressing different types of artifacts in the B-mode ultrasound images using a switchable network with adaptive instance normalization (AdaIN) layers \cite{Khan2022} has demonstrated its capability of transferring between different target domains by simply incorporating a AdaIN code in the network. Nevertheless, the network does not directly use the raw radio frequency (RF) ultrasound signals. Instead, the mapping was from the beamformed RF data to the image.

Enlightened by the aforementioned studies, in this work, we propose a deep beamformer for PAM based on generative adversarial network (GAN) which improves the performance of our previously proposed network for PAM \cite{yi2022}. The deep beamformer is capable of reconstructing high-quality PAM images with low computational cost directly using RF signals. It is also switchable for different transducer arrays to process RF signals with different characteristics. These arrays range from linear to phased arrays whose bandwidth covers 1--15 MHz. In the following sections, we first briefly recall the available beamforming methods for PAM, including TEA, EISRCB, RLPB and DAX-RCB (Sec. II). We then present the architecture of the proposed deep beamformer, the simulation and experimental datasets used to train and test the deep beamformer (Sec. III). The performance of the deep beamformer is evaluated by the comparison with the aforementioned available beamforming methods for PAM for three different transducer arrays (Sec. IV), which is followed by Discussion and Conclusions in Sec. V and Sec. VI, respectively.

\section{Related work}
The goal of PAM is to reconstruct a map representing the spatial distribution of the cavitation energy. The map can be formed based on the DAS operation. Considering an array with $N$  elements, the $j$th delayed element signal is given as:
\begin{equation}
s_j(\boldsymbol{r},t)=d_j(\boldsymbol{r})p_j(t+\tau_j(\boldsymbol{r}))
\label{DelayedSignal}
\end{equation}
where $p_j(t)$ is the passively received RF signal by the $j$th element, $\tau_j(\boldsymbol{r})$ and $d_j(\boldsymbol{r})$ are the traveling time and distance of the sound wave from location $\boldsymbol{r}$ to the $j$th element, respectively.
The delayed element signals are firstly weighted and summed across the receiving channels coherently. Then, the signal energy of the coherently summed signal over a period of time is calculated and scaled by a factor of $\frac{4\pi}{\rho_0 c}$ to represent the cavitation source energy at $\boldsymbol{r}$:
\begin{equation}
\Psi(\boldsymbol{r})=\frac{4\pi}{\rho_0 c}\int_{T_0}^{T_0+\triangle T}(\boldsymbol{w(\boldsymbol{r})}^T\boldsymbol{s}(\boldsymbol{r},t))^2 dt
\label{EnergyTEA}
\end{equation}
where $\rho_0$ is the density of the medium, $T_0$ is the onset of the interested time interval [$T_0$,$T_0+\triangle T$] for the RF signal, and $\boldsymbol{w(\boldsymbol{r})}$ is the channel specific weights for location $\boldsymbol{r}$.

Estimation of cavitation source energy depends on the choice of $\boldsymbol{w}$. TEA \cite{Norton2000} assumes uniform channel weights, whereas data-adaptive beamformers calculate the weights by minimizing the output power of interfering signals. Here, we briefly summarize the weight calculation for the data-adaptive beamfomers (Table~\ref{table:WeightCalcultation}). 

1) EISRCB\cite{Lu2018}: Considering steering vector uncertainty, RCB utilizes the covariance matrix of the delayed signals ($\boldsymbol{R}(\boldsymbol{r})=\int_{T_0}^{T_0+\triangle T} \boldsymbol{s}(\boldsymbol{r},t)\boldsymbol{s}(\boldsymbol{r},t)^T dt$) and minimizes the output power of interfering signals (Eq. (4)) to suppress artifacts \cite{Coviello2015}. To further improve the capability of artifacts suppression, EISRCB calculates the weights by projecting the RCB weights ($\boldsymbol{w}_{RCB}$) onto the signal subspace of $\boldsymbol{R}$ by Eq. (5).

2) DAX-RCB\cite{Lu2020}: DAX-RCB combines dual apodization with cross-correlation (DAX-TEA) \cite{Lu2019} that uses two complementary apodization functions to suppress unwanted sidelobes, clutter signals, and the artifacts in RCB in the region where the coherence is weak. The received signals are divided by a pair of complementary apodization functions which become two groups of signals that are used to calculate the minimum variance weights by Eq. (4) and then beamformed to obtained $\boldsymbol{s}_1(\boldsymbol{r},t)=\boldsymbol{w}^T_{RCB1}\boldsymbol{s}(\boldsymbol{r},t)$  and $\boldsymbol{s}_2(\boldsymbol{r},t)=\boldsymbol{w}^T_{RCB2}\boldsymbol{s}(\boldsymbol{r},t)$. The weights in DAX-RCB are then calculated according to Eq. (6), which utilizes the normalized cross-correlation coefficient to evaluate the similarity between $\boldsymbol{s}_1(\boldsymbol{r},t)$  and $\boldsymbol{s}_2(\boldsymbol{r},t)$.

3) RLPB\cite{Lyka2018}: The aforementioned data-adaptive beamformers assume that the signals of interest follow Gaussian distribution while it was argued that broadband bubble emissions may deviate from Gaussian distribution. RLPB thus utilizes higher order statistics of signals to optimize the weighting vector. The weights are obtained by minimizing the $l_\infty$-norm of the beamformer’s output according to Eq. (7). 

In summary, TEA, a non-data-adaptive beamformer, uses equal weighting factors for all the channels and has low image resolution and high-level artifacts. Data-adaptive beamformers including EISRCB, DAX-RCB and RLPB were proposed to improve image quality for PAM. These methods adopt various strategies to calculate the weights for image quality enhancement while some are sensitive to the required user-defined parameters. Also, their computational costs are different. TEA is the most computationally efficient method and RLPB is the most computationally expensive method. During the implementation of the data-adaptive beamformers for PAM, we found that EISRCB is less sensitive to the user-defined parameters and is less computationally intensive among the data-adaptive beamformers, while the reconstructed image quality is among the best. Therefore, we chose EISRCB to reconstruct PAM images for training the network.

\section{Methods}
\subsection{Simulated Data}
A simulated dataset of MB cavitation signals emitted from a single bubble cloud and two bubble clouds was built. The cavitation signals located at typical targeting locations in focused ultrasound (FUS) experiments acquired by three different transducer arrays (ATL P4-1, L7-4, CL15-7) were simulated (Table~\ref{table:TransducerParameters}). These arrays include linear array and phased array and their working bandwidth ranges low (1--4MHz), middle (4--7MHz) and high frequency (7--15MHz). The length of the recorded cavitation signals was 2048 time samples, which corresponds to a duration of 102 $\mu$s, 98 $\mu$s and 58 $\mu$s for the three arrays, respectively. 

\begin{figure}[!t]
\centerline{\includegraphics[width=0.45\columnwidth]{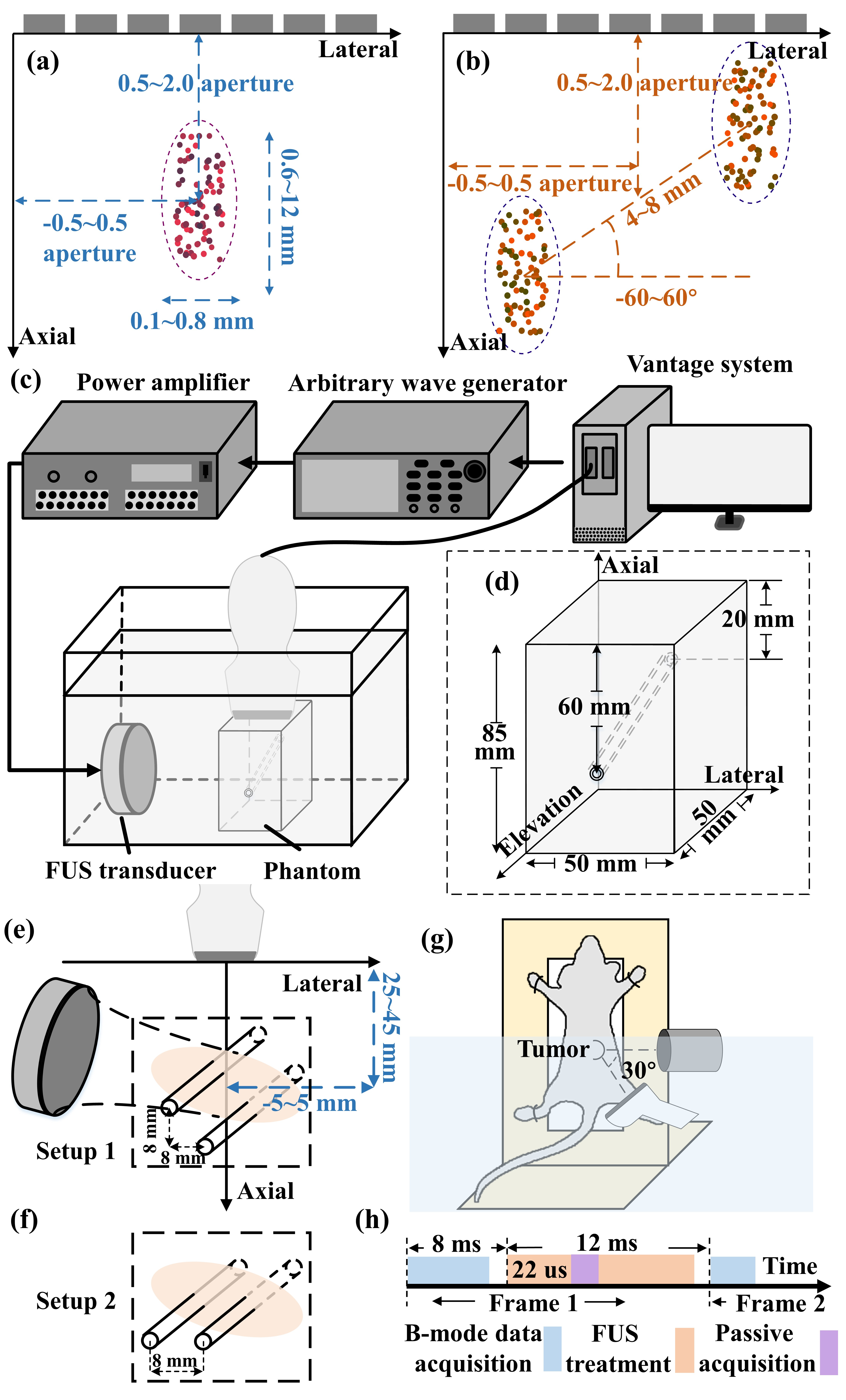}}
\caption{Distribution of (a) one bubble cloud and (b) two bubble clouds in the simulations. (c) Schematic diagram of the experimental setup. (d) The phantom with an inclined tube embedded. Experimental setup 1 (e) and 2 (f) of multiple bubble clouds for tubes. (g) Experimental setup of FUS treatment for mice and (h) the sequence for FUS treatment and data acquisition for B-mode imaging and PAM.}
\label{fig:Setup} 
\end{figure}

We assumed that the bubbles in the cloud (bubble count varied between 20--100 randomly) are randomly distributed in an elliptical area defined by the -3 dB area of the point spread function of a passive acoustic imaging system using a linear array~\cite{gyongy2010localization,gray2020}.
Thus, the size of the elliptical area depends on the imaging depth, and its major and minor axis ranged 0.6--12 mm and 0.1--0.8 mm, respectively (Fig.~\ref{fig:Setup}(a)). 
For single bubble cloud case, the center of the bubble cloud varied randomly (uniform distribution) within the full aperture range of the imaging array in the lateral direction and between 12.8--57.6 mm in the axial direction (F-number = 0.6--2.0) (Fig.~\ref{fig:Setup}(a)).
For two bubble clouds case, the center of the clouds were at both ends of a circle's diameter (randomly varied between 4 mm--8 mm).
The angle between the two ends and the center of the circle ranged from -60$^\circ$--60$^\circ$, respectively.
The center of the circle also varied randomly with the same settings as the ones for single bubble cloud's center. (Fig.~\ref{fig:Setup}(b)).

The Marmottant model \cite{Marmottant2005} and Vokurka model \cite{Karel2002} were used to generate the acoustic emissions of stable and inertial cavitation activities, respectively. The cavitation signals were generated such that the bubbles were excited by a driving sinusoidal signal whose frequency varied among 0.5 MHz, 1 MHz and 1.5 MHz for the single bubble cloud case, respectively.
For two bubble clouds case, the excitation frequency was 0.5 MHz.
The number of excitation cycles varied between 20--100 following uniform distribution. With the assumption that the bubbles do not interact with each other, we applied the Marmottant model to generate the cavitation signals for each bubble in the cloud, with randomly selected acoustic pressure of the excitation signal and bubble diameter. Specifically, to account for variations in bubble size and bubble-received acoustic pressure in experiments, the radius of the bubbles and the excitation pressure followed Gaussian distribution whose mean values were uniformly distributed (Table~\ref{table:SimulationParameters}). The cavitation signals received by the arrays were simulated using k-Wave \cite{Bradley2010}, with the input pressure sources defined by the location of the bubbles and the associated bubble acoustic emission signals. For the Vokurka model, the simulated cavitation signals were calculated by its analytic expression directly \cite{Karel2002,Lu2018}. In the expression, the acoustic pressure, phase offset and time constant of every excitation cycle were assumed normally distributed with the parameters listed in Table~\ref{table:SimulationParameters}. All the simulated cavitation signals were band-pass filtered to match the working bandwidth of the receiving arrays. The medium was homogeneous, but with the SoS varied between 1480--1600 m/s (uniform distribution) among the simulated acquisitions.

In total, the simulated dataset contains 42000 acquisitions, including 12000 and 2000 acquisitions with each transducer for one and two bubble clouds cases, respectively.

\begin{table}
\renewcommand\arraystretch{2}
\centering
\caption{Weight calculation in TEA, RCB, EISRCB, DAX-RCB, RLPB}
\label{table:WeightCalcultation}
\setlength{\tabcolsep}{5pt}
\begin{tabular}{l|l}
\hline
Methods & 
Weight calculation \\
\hline
TEA \cite{Norton2000} & 
$
\boldsymbol{w}(\boldsymbol{r})=\frac{1}{N}\boldsymbol{1}
$
~~~~~~~~~~~~~~~~~~~~~~~~~~~~~~~~~~~~~\;\;\;\;\;\, (3)
\\
\hline
RCB \cite{Coviello2015} & 
\makecell[l]{
$
\min_{\boldsymbol{a}}\boldsymbol{a}^T\boldsymbol{R}(\boldsymbol{r})^{-1}\boldsymbol{a} \ s.t.\| \boldsymbol{a}-\overline{\boldsymbol{a}} \|^2 \leq \varepsilon
\hspace{\fill}
$ (4)
\\ $\boldsymbol{a}$, $\boldsymbol{\bar a}$ is the actual and assumed steering \\
vector, $\varepsilon$ is a user-defined parameter~~~~~~~~~~~\;\;\;\;\,} 
\\
\hline
EISRCB \cite{Lu2018} & 
\makecell[l]{
$
\boldsymbol{w(\boldsymbol{r})}=\boldsymbol{U}_s\boldsymbol{U}_s^T\boldsymbol{w(\boldsymbol{r})}_{RCB}, s.t. \gamma_1\leq\delta\gamma_L 
\hspace{\fill}
$ (5)
\\ $\boldsymbol{w}_{RCB}$ is the weight of RCB
\\ $\boldsymbol{U}_s$ comprises the $L$ maximum eigenvalues \\
of $\boldsymbol{R}$, $\gamma_1$, $\gamma_L$ are the biggest and smallest \\ 
eigenvalues~~~~~~~~~~~~~~~~~~~~~~~~~~~~~~~~~~~~~~~~~~~~~~~~~} 
\\
\hline
DAX-RCB \cite{Lu2020} & 
\makecell[l]{ 
$
\boldsymbol{w(\boldsymbol{r})}=\frac{Cov(\boldsymbol{s}_1(\boldsymbol{r},t),\boldsymbol{s}_2(\boldsymbol{r},t))}{\sqrt{D(\boldsymbol{s}_1(\boldsymbol{r},t))}\sqrt{D(\boldsymbol{s}_2(\boldsymbol{r},t))}}
\hspace{\fill}
$ (6)
\\
$Cov$, $D$ are the covariance and variance \\ 
functions, respectively ~~~~~~~~~~~~~~~~~~~~~~~~~~~~~~\;\;\,
} 
\\
\hline
RLPB \cite{Lyka2018} & 
\makecell[l]{
$
\min_{\boldsymbol{w}} \| \boldsymbol{s}^T\boldsymbol{w(\boldsymbol{r})} \|_\infty \ s.t.\boldsymbol{a}^T\boldsymbol{w}-\tau\|\boldsymbol{w}\|_\infty \geq 1 
\hspace{\fill}
$ (7)
\\
$\tau$ is a user-defined parameter ~~~~~~~~~~~~~~~~~~~\;\;\;\;\,
}
\\
\hline

\end{tabular}
\label{tab1}
\end{table}

\subsection{Experimental Data}
\begin{table}
\centering
\caption{Transducer parameters}
\label{table:TransducerParameters}
\setlength{\tabcolsep}{3pt}
\begin{tabular}{c|c|c|c|c}
\hline
Transducer& 
Elements& 
Aperture &
\makecell[c]{Center \\ frequency} &
Bandwith \\
\hline
P4-1 & 96 & 28.8 mm & 2.5 MHz & 80\% \\
L7-4 & 128 & 38.4 mm & 5 MHz & 60\% \\
CL15-7 & 128 & 25.6 mm & 9 MHz & 60\% \\
\hline
\end{tabular}
\label{tab1}
\end{table}

In the experimental dataset, three different transducer arrays (P4-1, L7-4, CL15-7) were successively interfaced to a Vantage 256 system (Verasonics, Kirkland, WA, USA) to receive the MB cavitation signals. To initiate the MB cavitation signals, two FUS transducers of different center frequencies (Doppler, Guangzhou, China) were used for \emph{in vitro} and \emph{in vivo} experiments, respectively. The experiments were performed in deionized and degased water.
The recorded cavitation signals have 2048 time samples, which corresponds to a duration of 102 $\mu$s, 98 $\mu$s and 58 $\mu$s for the three arrays, respectively.

For the \emph{in vitro} experiments of imaging a single bubble cloud, phantoms with different SoS (1480 m/s, 1490 m/s, 1516 m/s, 1560 m/s) were prepared by mixing 2\% (w/v) agar powder and 3--12\% (v/v) ethanol in deionized and degassed water. A silicone tube (0.8 mm inner diameter) was inclined in the phantom, whose two ends were 20 mm and 60 mm far from the top of the phantom, respectively (Fig.~\ref{fig:Setup}(c)). The tube was filled with MB solution (1$\times10^7$ MBs/ml, Vevo MicroMarker, Visualsonics, Canada) flowing at a constant speed (1 ml/min). The array was moved over the phantom so that the imaged location of cavitation source varied between -10--10 mm away from the center of the imaging array in the lateral direction, and between 25--55 mm in the axial direction. The FUS transducer with full width half maximum (FWHM) of 3.1 mm (lateral) and 21.9 mm (axial) (center frequency: 0.5 MHz, focal depth: 64 mm, aperture diameter: 64 mm) was driven at various pulse lengths to initiate MB cavitation activity in the phantoms. With P4-1, 8000 acquisitions were made on the phantoms of the four different SoS (2000 acquisitions per SoS value), and the number of excitation cycles was 120. With L7-4, 2000 acquisitions were made on the phantom of 1516 m/s SoS, and the number of excitation cycles was 120. With CL15-7, 6000 acquisitions were made on the phantom of 1516 m/s SoS, and the number of excitation cycles varied among 20, 30 and 50. 

For imaging two bubble clouds, two setups of two parallel tubes (0.8 mm inner diameter) were prepared.
In one setup, the two tubes were positioned with 8 mm separation vertically and horizontally  (Fig.~\ref{fig:Setup}(e)). In another, the two tubes were aligned 8 mm apart horizontally (Fig.~\ref{fig:Setup}(f)). 
The tubes were positioned below the arrays such that their cross-sections varied between -5--5 mm in the lateral direction and between 25--45 mm in the axial direction (Fig.~\ref{fig:Setup}(e)).  
The tubes were filled with MB solutions (1$\times10^7$ MBs/ml, Sonovue, Bracco, Italy) flowing at a speed (1 ml/min). 
The 0.5 MHz FUS transducer (20--50 excitation cycles) was tilted at a certain angle in the horizontal direction to initiate MB cavitation activity in the tubes. For each array, 800 acquisition were acquired and the SoS of the water was estimated as 1485 m/s.

\begin{table}
\centering
\caption{Simulation parameters}
\label{table:SimulationParameters}
\setlength{\tabcolsep}{3pt}
\begin{tabular}{c|c|c}
\hline
& 
Parameter & 
Value \\
\hline
\multirow{3}{*}{\makecell[c]{Vokurka \\ model}} & acoustic pressure & $\boldsymbol{N}$(14 MPa, 10 kPa)  \\
\multirow{3}{*}{} & phase offset & $\boldsymbol{N}$(14 $\mu$s, 10 ns) \\
\multirow{3}{*}{} & time constant & $\boldsymbol{N}$(2 $\mu$s, 0.5 ns) \\
\hline
\multirow{2}{*}[-1ex]{\makecell[c]{Marmottant \\ model}} & acoustic pressure & \makecell[c]{$\boldsymbol{N}$($\mu_a$, 10 kPa), \\ $\mu_a \in$ $\boldsymbol{U}$(0.1 MPa, 1 MPa)} \\
\multirow{2}{*}{} & bubble radius & \makecell[c]{$\boldsymbol{N}$($\mu_b$, 0.1 $\mu$m), \\ $\mu_b \in$ $\boldsymbol{U}$(1 $\mu$m, 3 $\mu$m)} \\
\hline
\end{tabular}
\label{tab1}
\end{table}

For the \emph{in vivo} experiments, Hepa1-6 liver cancer cells (10$^7$ cells, 50 $\mu$L) were subcutaneously injected into two C57BL/6 mice (male, 5--6 weeks old, weighing 20--22g), and allowed to form tumors sized 200--400 mm$^3$. During FUS treatments, the mice, anesthetized by tribromoethanol (30 $\mu$l/g), was tied to a shelf and placed in the water. The CL15-7 array was facing the tumor, and the FUS transducer with FWHM of 1.5 mm (lateral) and 10.7 mm (axial) (center frequency: 1 MHz, focal depth: 24 mm, aperture diameter: 24 mm) was placed in the same horizontal plane as the CL15-7 array (Fig.~\ref{fig:Setup}(d)). A bolus of 50 $\mu$L MB solution (Sonazoid, GE Healthcare, Oslo, Norway) was injected into the tumor which was then excited by the FUS transducer for 100 cycles. The cavitation signals (8000 acquisitions) were acquired by the CL15-7 array for a duration of 58 $\mu$s and sampled at 35.6 MHz. The B-mode images of the tumor were also acquired using plane wave imaging (15 plane waves, -20$^\circ$--20$^\circ$) with the same array. The imaging sequence and FUS treatment sequence were interleaved and the effective data acquisition rate was 50 Hz (Fig.~\ref{fig:Setup}(e)). All animal experiments were approved by the Institutional Animal Care and Use Committee of the ShanghaiTech University (Protocol 20220516001). 

The peak negative pressure of the FUS transducers measured using a hydrophone (NH0200, Precision Acoustics, Dorchester, U.K.), was between 0.5-1 MPa for the \emph{in vivo} experiments, and was between 0.75-1 MPa and 1-1.5 MPa for the \emph{in vitro} single bubble cloud and multiple bubble clouds experiments, respectively.

\subsection{Image Reconstruction}

The PAM images in the dataset were reconstructed by the EISRCB which were used as the ground-truth images. The parameters used in the EISRCB were chosen as $\delta$ = 0.5, $\varepsilon$ = 20 and 30 for single and multiple sources, respectively, after trials \cite{Lu2018}. The reconstruction parameters for different transducer models are listed in Table~\ref{table:ImageReconstructionParameters}. All the PAM images have the same size (512$\times$512) to facilitate network training. The pixel size in the lateral direction was 0.25 times of the pitch of each transducer model. In the axial direction, it was adjusted to be less than a half of the wavelength of the array’s center frequency. The center of the arrays’ aperture was placed at (0, 0) mm in both directions. The lateral image range covered the full aperture of the respective transducer models array. The axial image range was 0--100 mm for P4-1, 15--65 mm for L7-4, 10--50 mm for CL15-7 to balance the image size and resolution. All image reconstructions were performed in Matlab (R2016B, MathWorks, Inc., Natick, MA, USA) on a workstation (Intel Xeon 3 GHz, 128 GB memory, NVIDIA GeForce RTX 4090). 

\subsection{Input Data}

To unify the dimension of the input RF signals received by different arrays, all signals were firstly padded with zeros to a size of 2048$\times$128. The unified data was then resized to $\boldsymbol{Y} \in \mathbb{R}^{256 \times 128}$. To introduce a \emph{priori} information about the transducer models, a one-hot code mask vector $\boldsymbol{m} \in \mathbb{R}^{1 \times 3}$ \cite{choi2018} was created for each transducer: [1,0,0] for P4-1, [0,1,0] for L7-4, [0,0,1] for CL15-7. This mask vector $\boldsymbol{m}$ was expanded to a three-dimensional matrix $\boldsymbol{M} \in \mathbb{R}^{256 \times 128 \times 3}$. Then, $\boldsymbol{M}$ was stacked with $\boldsymbol{Y}$ to generate stacked data $\boldsymbol{Z} \in \mathbb{R}^{256 \times 128 \times 4}$ that served as the input of the network.

\subsection{Network Architecture}
The proposed switchable deep beamformer is based on the pix2pixGAN \cite{Isola2017} and our previously proposed neural network \cite{yi2022}, which consists of a generator G, a discriminator D and a classifier K.

\begin{figure*}[!t]
\centerline{\includegraphics[width=1.0\columnwidth]{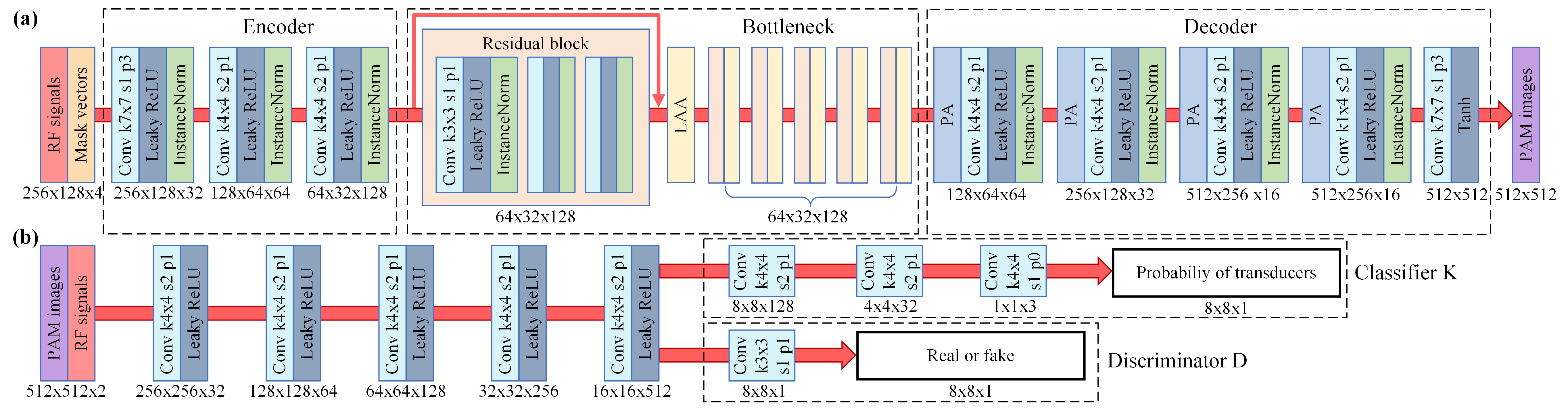}}
\caption{Architecture of the switchable deep beamformer for PAM: (a) the generator, (b) the discriminator/classifier. The dimension below each blocks denotes the size of the output feature maps. k, s, p denotes the number of kernel size, stride and size of padding, respectively. LAA: local aware attention; PA: pixel attention.}
\label{fig:Network} 
\end{figure*}

The generator G implemented with an encoder-decoder structure (Fig.~\ref{fig:Network}(a)) was trained to reconstruct high-quality PAM images from RF signals for different transducer arrays (P4-1, L7-4, CL15-7). Different from a traditional encoder-decoder structure, we designed the generator G with an asymmetric structure which allows asymmetric input and output size. In the encoder, there are three levels of convolutional layers with instance normalization and leaky rectified linear unit (ReLU) activation. 
The encoder transforms the input data from 4 channels to the feature maps with 128 channels. There is a bottleneck structure formed by cascading 6 residual blocks between the encoder and the decoder. Each residual block is composed of 3 convolutional layers with leaky ReLU activation and instance normalization. In the decoder, the first four levels consist of convolutional layers with instance normalization and leaky ReLU activation, transforming the feature maps from 128 channels to 16 channels. The last level of the decoder, which is a convolutional layer with hyperbolic tangent (Tanh) activation finally merges all the feature maps into one image. To improve the performance of the network, two attention blocks, local aware attention \cite{guo2019} and pixel attention \cite{zhao2020}, were added. Local aware attention, which enhances the ability of extracting high-frequency features, is performed after each residual block in the bottleneck. Pixel attention is performed on each convolutional layer in the decoder to improve the global attention of the generator G.

The discriminator D and the classifier K share the same network comprised of 6 levels of convolutional layers with leaky ReLU activation and a separate convolutional layer without activation (Fig.~\ref{fig:Network}(b)). The discriminator D was trained to differentiate between the true images reconstructed by EISRCB and the fake images reconstructed by the generator G. The output of the discriminator D is an 8$\times$8 boolean matrix in which the value (1 or 0) of the matrix element represents whether the corresponding 64 patches assembling the image is real or fake. This approach has been found to enhance the high-frequency components of the output images. The classifier K outputs probability of the type of transducer model, i.e. to which transducer array an image corresponds. This helps the generator G learn better the prior information carried by the mask vector $\boldsymbol{m}$. For training stability, PAM images and the corresponding RF signals were stacked and used as the inputs to the discriminator D and the classifier K.

\begin{table}
\centering
\caption{Image reconstruction parameters}
\label{table:ImageReconstructionParameters}
\setlength{\tabcolsep}{3pt}
\begin{tabular}{c|c|c|c}
\hline
Transducer& 
Size& 
\makecell[c]{Imaging range \\ (mm $\times$ mm)}&
\makecell[c]{Pixel size \\ (mm $\times$ mm)}  \\
\hline
P4-1 & 512 $\times$ 512 & 102.4 $\times$ 38.4 & 0.2 $\times$ 0.075 \\
L7-4 & 512 $\times$ 512 & 51.2 $\times$ 38.4 & 0.1 $\times$ 0.075 \\
CL15-7 & 512 $\times$ 512 & 41.0 $\times$ 25.6 & 0.08 $\times$ 0.050 \\
\hline
\end{tabular}
\end{table}

\subsection{Network Training}

Traditional GAN adopts the cross-entropy loss to minimize the distance between original images and target images. However, the cross-entropy loss was found unsuitable for the scenario of directly mapping RF signals to images. Therefore, we propose an improved loss scheme based on the training scheme of Wasserstein GAN \cite{gulrajani2017}. The new loss $\mathcal{L}$  consists of adversarial loss $\mathcal{L}_{adv}$(G,D), reconstruction loss $\mathcal{L}_{rec}$(G), dice loss $\mathcal{L}_{dic}$(G), gradient penalty $\mathcal{L}_{gp}$(D) and classification loss $\mathcal{L}_{cls}$(K):
\setcounter{equation}{7}
\begin{equation}
\begin{aligned}
\mathcal{L} &= \\
&\lambda_{adv}\mathcal{L}_{adv}(G,D) + \lambda_{rec}\mathcal{L}_{rec}(G) + \lambda_{dic}\mathcal{L}_{dic}(G) \\
&+ \lambda_{cls}\mathcal{L}_{cls}(K) + \lambda_{gp}\mathcal{L}_{gp}(D)
\end{aligned}
\end{equation}
where $\lambda_{adv}$, $\lambda_{rec}$, $\lambda_{dic}$, $\lambda_{cls}$ and $\lambda_{gp}$ are the weighting parameters.

1) Adversarial Loss: It measures the statistical distance quantifying the similarity between the distribution of the deep beamformer yielded images and the distribution of the ground truth images, which is defined as:
\begin{equation}
\mathcal{L}_{adv}(G,D) = \mathbb{E}[D(G(\boldsymbol{x}))] - \mathbb{E}[D(\boldsymbol{y})]
\end{equation}
where $\boldsymbol{x}$ is the RF signals and $\boldsymbol{y}$ is the ground truth image. $\mathbb{E}$ is the mathematical expectation function. 

2) Reconstruction Loss: To speed up the training, mean squared error between the network generated image and the ground truth image is calculated as:
\begin{equation}
\mathcal{L}_{rec}(G) = \mathbb{E}[\|\boldsymbol{y}-G(\boldsymbol{x})\|_2^2]
\end{equation}

3) Dice Loss: Under the reconstruction loss constraint, the generated image starts with low-energy and smooth characteristics and gradually approaches the ground truth during training. However, due to the smaller number of high-energy pixels compared to low-energy pixels in PAM images, the discrepancy between the high-energy regions in the generated images and the ground truth can only result in a sufficiently small reconstruction loss. This makes the network difficult to generate PAM images with the maximum values of 1. To address this issue, dice similarity coefficient \cite{milletari2016}, which encourages the network to output values closer to 1 in regions with non-zero values, was utilized:
\begin{equation}
\mathcal{L}_{dic}(G) = 1 - \frac{2\sum_{i=1}^N y_i G(x)_i}{\sum_{i=1}^N y_i^2 + \sum_{i=1}^N G(x)_i^2}
\label{diceloss}
\end{equation}
where $i$ indexes the $i$th pixel, and $N$ is the number of image pixels.

4) Classification Loss: The classifier K was trained to predict the same discrete probability distribution as the ones in the ground truth. Kullback-Leiber divergence was selected to represent the loss:
\begin{equation}
\mathcal{L}_{cls}(K) = \mathbb{E}[-\log K(\boldsymbol{y})]
\end{equation}

5) Gradient Penalty: Imposing the condition of 1-Lipschitz \cite{gulrajani2017} on the discriminator D improves the training process without compromising the stability of the training:
\begin{equation}
\mathcal{L}_{gp}(D) = -\mathbb{E}[\max(0,\|\nabla_{\hat{\boldsymbol{y}}}D(\boldsymbol{y}) \|-1)^2]
\end{equation}
where $\hat{\boldsymbol{y}}=\alpha\boldsymbol{y}+(1-\alpha)G(\boldsymbol{x})$ and $\alpha$ is a random value following uniform distribution between [0,1].

The proposed network was trained using both simulated and experimental data with a training-to-test set ratio of 7:1. 
During the training, white noise was added to the input RF signals for every acquisition to randomly vary the signal-to-noise ratio (SNR) within 0--15 dB (uniform distribution), covering the reported SNR range in previous works \cite{gray2020,Patel2019}.
The network was optimized with the Adam \cite{kingma2014adam}, starting with a learning rate of $5e^{-5}$ and linearly decreases to $5e^{-6}$. The total epoch was 128, and the batch size was 64. The weights of the loss function, $\lambda_{adv}$, $\lambda_{rec}$, $\lambda_{dic}$, $\lambda_{cls}$ and $\lambda_{gp}$ were set to 1, 100, 1, 0.1 and 30, respectively. The network was implemented on the Python with PyTorch and trained with the RTX 4090 GPU.

\subsection{Evaluation Metrics}
The proposed deep beamformer was compared with the EISRCB, RLPB, DAX-RCB and TEA using the simulated and experimental data in terms of the energy spread area, artifacts suppression capability and computational complexity. The energy spread area was evaluate as the $A_{-3dB}$ area where the pixel values were greater than half of the maximum value in the image \cite{Lu2019,Lu2020}. Image SNR (ISNR) was used to evaluate the artifacts suppression capability, defined as:
\begin{equation}
ISNR = 10\log_{10}(\Psi_{Inside}/\Psi_{Outside})
\end{equation}
where $\Psi_{Inside}$ is the mean pixel value inside $A_{-3dB}$ area, $\Psi_{Outside}$ is the mean pixel value outside $A_{-3dB}$ area but inside $\tilde{A}_{-20dB}$ area, the union of -20 dB area of the images reconstructed by different methods. In addition, floating-point operations (FLOPs) and running time required to reconstruct an image was used to evaluate the computational complexity of the deep beamformer and the other methods. The FLOPs for the deep beamformer was estimated using the PyTorch library THOP \cite{thop}, while for other beamformers, the FLOPs was estimated using the Matlab function FLOPS \cite{FLOPS}. 

To evaluate the capability of localizing cavitation sources with the deep beamformer, we evaluated source position deviation between the images reconstructed by the deep beamformer and EISRCB using the test dataset, with
the source position defined as the centroid $\boldsymbol{r}_c$ in the $A_{-3dB}$  area:
\begin{equation}
\boldsymbol{r}_c = \frac{\sum_{i=1}^N \Psi(\boldsymbol{r}_i) \boldsymbol{r}_i}{\sum_{i=1}^N \Psi(\boldsymbol{r}_i)} \ s.t. \boldsymbol{r}_i \in A_{-3dB}
\end{equation}
and the deviation defined as Euclidean distance between the centroids. 
To visualize the difference, the cumulative probability distribution of the deviation by wavelength (defined at the center frequency of the corresponding transducer array) was also evaluated.

To compare different data-adaptive beamformers, we used randomly selected 120 test data for each transducer model to calculate the evaluation metrics, due to the limitation that they require too much time, particularly the RLPB method, to generate an image for the settings used in this work (see Sec.~\ref{sec:ComputationalComplexity}). To evaluate the deep beamformer, all test data were used to calculate the evaluation metrics.

\section{Results}

\subsection{Comparison of the data adaptive beamformers}
\begin{figure}[!t]
\centerline{\includegraphics[width=0.55\columnwidth]{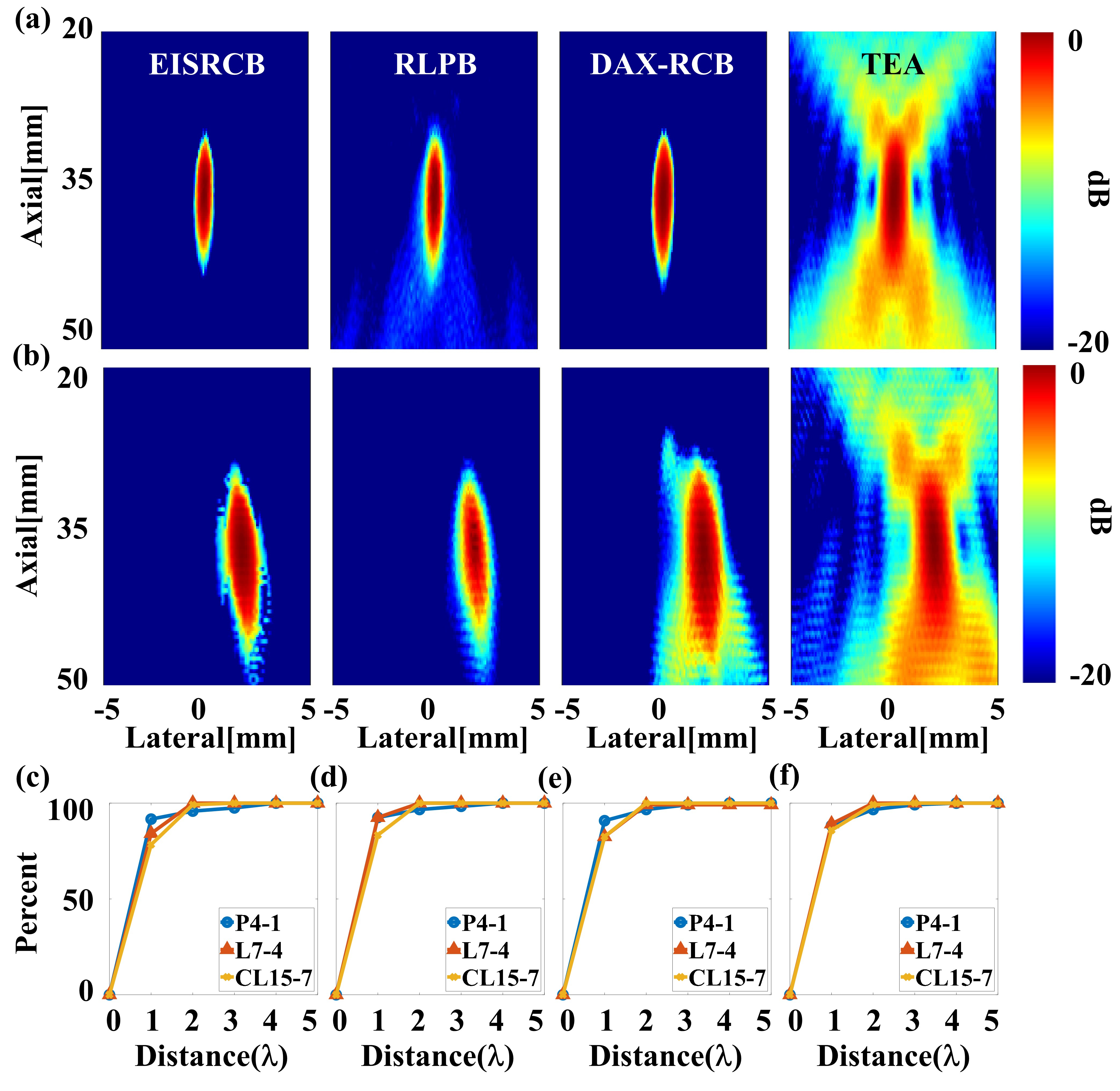}}
\caption{(a) Simulated and (b) \emph{in vitro} experimental PAM images reconstructed by the EISRCB, RLPB, DAX-RCB and TEA for P4-1. Cumulative probability distribution of the Euclidean distance between the center of the bubble clouds and source position localized by (c) EISRCB, (d) RLPB, (e) DAX-RCB and (f) TEA.}
\label{fig:AdaptiveResults} 
\end{figure}

In our simulated and experimental data of imaging single bubble cloud, the RLPB method achieved the optimal energy spread area $A_{-3dB}$, showing 55.6--75.0\% improvements in average compared with the TEA method (Fig.~\ref{fig:AdaptiveResults}, Table.~\ref{table:SingleSourcePerformance}).
In terms of the artifacts suppression capability, the EISRCB method was the best which improved the ISNR by 14.8--24.1 dB in average compared to the TEA method.
The difference in the average value of $A_{-3dB}$ between the EISRCB and RLPB images were less than 0.3 $mm^2$, except for the experimental data measured by P4-1 (4.7 $mm^2$) (Table.~\ref{table:SingleSourcePerformance}), indicating that the size of energy beam provided by EISRCB was similar to that by RLPB.
Moreover, the proportion of the Euclidean distance below 4 wavelengths between the center of the bubble cloud and the position localized by EISRCB was 100.0\% for all the transducer arrays, which was similar to the results of other beamformers (Fig.~\ref{fig:AdaptiveResults}(c--f)).
However, the computational time of EISRCB was about one order of magnitude faster than that of RLPB (see Sec.~\ref{sec:ComputationalComplexity}).
These results showed that EISRCB balanced well between image quality and computational time, and can localize cavitation sources accurately.
Therefore, EISRCB was chosen to establish the dataset for training the deep beamformer.

\subsection{Evaluation of the Deep Beamformer}
We used the simulated dataset to quantitatively evaluate the performance of the deep beamformer. 
\subsubsection{Image quality}
\begin{figure*}[!t]
\centerline{\includegraphics[width=1.0\columnwidth]{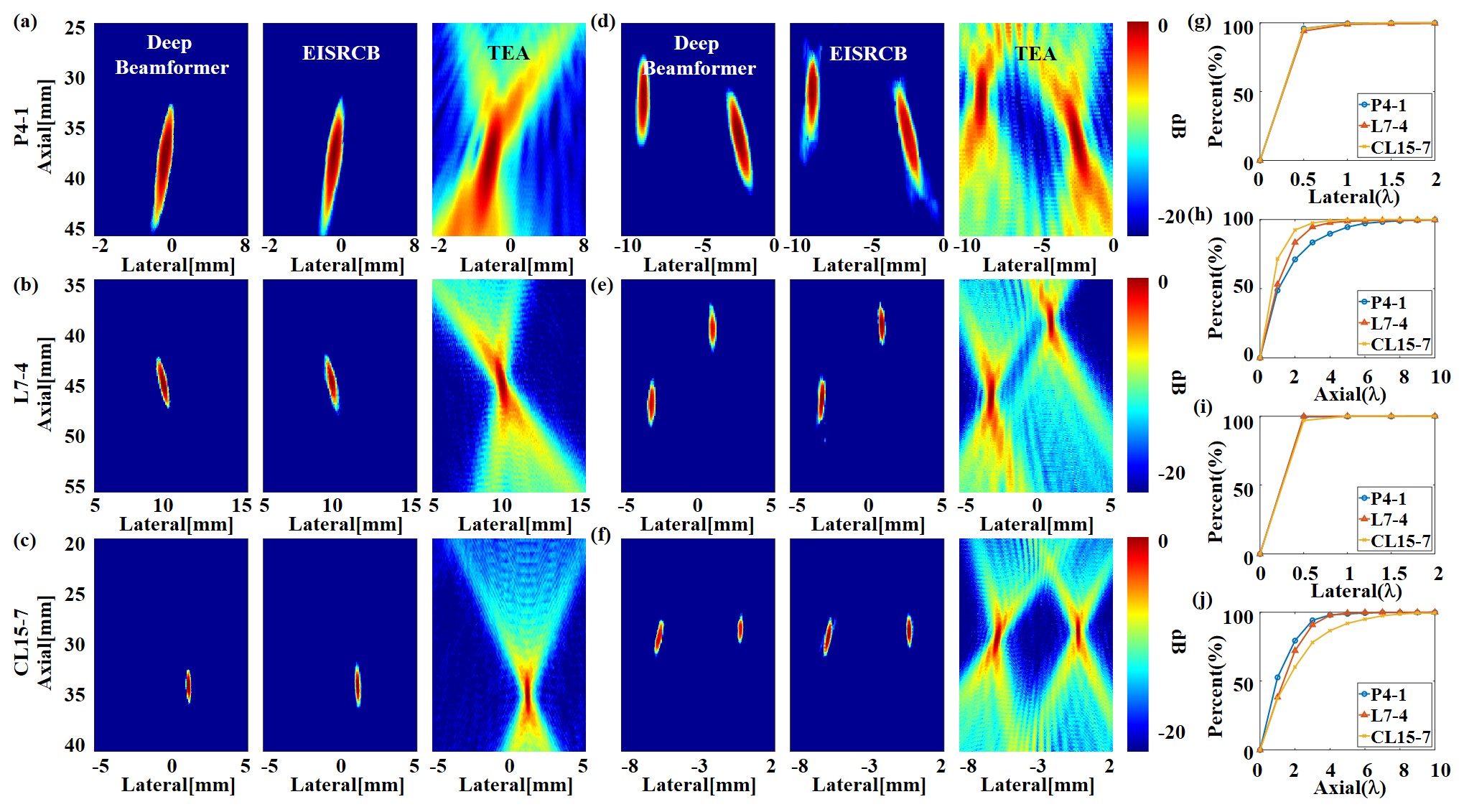}}
\caption{PAM images reconstructed by the deep beamformer, EISRCB, and TEA using the simulated test set of (a--c) single and (d--f) two bubble clouds for P4-1, L7-4 and CL15-7. Cumulative probability distribution of lateral position deviation and axial position deviation between the deep beamformer yielded images and EISRCB images for (g--h) single source and (i--j) multiple sources. The wavelength ($\lambda$) was evaluated at the center frequency of each transducer array.}
\label{fig:SimulatedResults} 
\end{figure*}

The high-intensity region was well localized in the images reconstructed by the deep beamformer, which showed good consistency with the ones by EISRCB for all the transducer arrays (Fig.~\ref{fig:SimulatedResults}(a--f)). 
For single cavitation sources, the difference in the average value of $A_{-3dB}$ and ISNR was 0.0--0.2 $mm^2$ and 0.0--1.2 dB in all the arrays (Table~\ref{table:SingleSourcePerformance}). Compared with TEA, the average $A_{-3dB}$ area of the PAM images reconstructed by the deep beamformer was reduced by about 55.7\%, 44.4\% and 62.5\% for P4-1, L7-4 and CL15-7, respectively (Table~\ref{table:SingleSourcePerformance}). For the average ISNR of the PAM images, it was improved by 20.0 dB, 22.3 dB and 22.9 dB by the deep beamformer for P4-1, L7-4 and CL15-7, respectively, compared with TEA images (Table~\ref{table:SingleSourcePerformance}). 
In the case of multiple cavitation sources, the difference in the average value of $A_{-3dB}$ area and ISNR was 0.2--0.7 $mm^2$ and 1.4--6.0 dB in all the arrays (Table~\ref{table:TwoSourcePerformance}). 
Compared to TEA, the average $A_{-3dB}$ area of the PAM images reconstructed by the deep beamformer was reduced by about 18.9\%, 29.4\% and 30.0\% for P4-1, L7-4 and CL15-7, respectively (Table~\ref{table:TwoSourcePerformance}). For the average ISNR of the PAM images, it was improved by 17.5 dB, 9.3 dB and 12.2 dB by the deep beamformer for P4-1, L7-4 and CL15-7, respectively, compared with TEA images (Table~\ref{table:TwoSourcePerformance}). 
These results demonstrated that the deep beamformer can reduce the size of energy beam and have strong artifacts suppression for both single and multiple cavitation sources.

\subsubsection{Source localization accuracy}
Compared to the EISRCB images, we observed negligible ($<$ 4 wavelengths) deviation of the localized source position in the deep beamformer images (Fig.~\ref{fig:SimulatedResults}(g--j)). For single cavitation source, the proportion of the deviation below 1 wavelengths was 97.8\%--99.2\% in the lateral direction, while the proportion of the deviation below 4 wavelengths was 89.7\%--99.2\% in the axial direction. 
For multiple cavitation sources, the percentage of the deviation below 1 wavelength was 98.2\%--99.8\% in the lateral direction, while the proportion of the deviation below 4 wavelengths was 86.6\%--97.9\% in the axial direction. 

\begin{table*}
\centering
\caption{Energy spread area $A_{-3dB}$ (mean $\pm$ standard deviation) and ISNR (mean $\pm$ standard deviation) of the simulated and \emph{in vitro} experimental PAM Images of single source, reconstructed by the deep beamformer, EISRCB, RLPB, DAX-RCB and TEA for different transducer arrays.}
\label{table:SingleSourcePerformance}
\setlength{\tabcolsep}{3pt}
\begin{tabular}{c|c|c|c|c|c|c|c}
\hline
\multirow{2}{*}{Data Type} & \multirow{2}{*}{Algorithm} & \multicolumn{3}{c|}{$A_{-3dB}$ ($mm^{2}$)} & \multicolumn{3}{c}{ISNR values (dB)} \\
\cline{3-8}
\multirow{2}{*}{} & \multirow{2}{*}{} 
& P4-1 & L7-4 & CL15-7 
& P4-1 & L7-4 & CL15-7\\
\hline
\multirow{5}{*}{\makecell[c]{simulated \\ data}} & deep beamformer 
& 3.5 $\pm$ 3.2 & 0.5 $\pm$ 0.3 & 0.3 $\pm$ 0.2
& 31.4 $\pm$ 4.2 & 36.1 $\pm$ 3.7 & 36.3 $\pm$ 2.9 \\
\multirow{5}{*}{} & EISRCB 
& 3.3 $\pm$ 3.4  & 0.5 $\pm$ 0.3 & 0.3 $\pm$ 0.2 
& 30.9 $\pm$ 3.7 & 36.1 $\pm$ 2.5 & 37.5 $\pm$ 2.6 \\
\multirow{5}{*}{} & RLPB 
& 3.0 $\pm$ 3.2 & 0.4 $\pm$ 0.3 & 0.3 $\pm$ 0.2
& 24.9 $\pm$ 5.0 & 29.3 $\pm$ 4.0 & 30.8 $\pm$ 2.7 \\
\multirow{5}{*}{} & DAX-RCB 
& 4.0 $\pm$ 3.8 & 0.6 $\pm$ 0.3 & 0.4 $\pm$ 0.3
& 28.9 $\pm$ 3.3 & 33.4 $\pm$ 2.2 & 34.8 $\pm$ 2.3 \\
\multirow{5}{*}{} & TEA
& 7.9 $\pm$ 8.9 & 0.9 $\pm$ 0.6 & 0.8 $\pm$ 0.6
& 11.4 $\pm$ 1.7 & 13.8 $\pm$ 0.7 & 13.4 $\pm$ 0.5 \\
\hline
\multirow{5}{*}{\makecell[c]{experimental \\ data}} & deep beamformer 
& 10.5 $\pm$ 5.1 & 0.7 $\pm$ 0.2 & 1.1 $\pm$ 0.7
& 26.4 $\pm$ 2.2 & 29.8 $\pm$ 1.8 & 28.5 $\pm$ 3.4 \\
\multirow{5}{*}{} & EISRCB 
& 10.5 $\pm$ 5.0 & 0.7 $\pm$ 0.3 & 1.1 $\pm$ 0.6
& 25.5 $\pm$ 2.4 & 27.9 $\pm$ 1.8 & 25.1 $\pm$ 3.1 \\
\multirow{5}{*}{} & RLPB 
& 5.8 $\pm$ 3.6 & 0.5 $\pm$ 0.4 & 0.8 $\pm$ 0.7
& 23.8 $\pm$ 3.1 & 23.9 $\pm$ 1.1 & 23.0 $\pm$ 2.8 \\
\multirow{5}{*}{} & DAX-RCB 
& 16.0 $\pm$ 7.4 & 1.4 $\pm$ 0.5 & 1.9 $\pm$ 1.2
& 20.2 $\pm$ 2.4 & 24.8 $\pm$ 1.7 & 23.6 $\pm$ 2.8 \\
\multirow{5}{*}{} & TEA
& 16.3 $\pm$ 7.8 & 2.0 $\pm$ 0.8 & 2.6 $\pm$ 2.1
& 9.3 $\pm$ 0.4 & 10.8 $\pm$ 0.7 & 10.3 $\pm$ 1.3 \\
\hline
\end{tabular}
\end{table*}

\subsubsection{Different noise levels}
We also explored the anti-noise performance of the deep beamformer, with the SNR ranging from 0 to 15 dB in the cavitation signals (Fig.~\ref{fig:SimuNoiseTest}).
The added noise was white noise, as commonly used in other studies to simulate the experimental conditions in cavitation imaging~\cite{Coviello2015,Lu2020,gray2020}.
In this range, the average ISNR of the deep beamformer and EISRCB images was 
30.8--35.5 dB (31.5--36.4 dB) and 22.9--32.2 dB (28.2--36.1 dB), respectively, for all arrays. 
In terms of source localization accuracy, the proportion of the deviation below 1 wavelength in the lateral direction negligibly decreased from 98.7\%--99.8\% (15 dB) to 98.2\%--99.5\% (0 dB), while the proportion of the deviation below 4 wavelengths decreased from 89.3\%--99.3\% (15 dB) to 88.1\%--98.3\% (0 dB) in the axial direction.
In the investigated noise levels (Fig.~\ref{fig:SimuNoiseTest}), the image quality and source localization accuracy of the deep beamformer was negligibly affected demonstrating its robust anti-noise capability against Gaussian noise.

\subsubsection{Computational Complexity}
\label{sec:ComputationalComplexity}
The FLOPs required by the deep beamformer, EISRCB, RLPB, DAX-RCB and TEA to reconstruct one image of 262144 pixels was  $1.1 \times 10^{10}$, $2.3 \times 10^{13}$, $4.2 \times 10^{14}$, $1.0 \times 10^{13}$, and $4.1 \times 10^{11}$, respectively.
The actual execution time of the beamformers depends on the implementation and the computation platform. For a fair comparison, we estimated the computation time for different beamformers on a single CPU thread. The average computational time required by the deep beamformer, EISRCB, RLPB, DAX-RCB and TEA to reconstruct one image of 262144 pixels was 0.24 s, 3112 s, 28835 s, 2627 s, 1246 s. 
We also compared the computation time of the deep beamformer and TEA on the GPU.
In average, the deep beamformer and TEA required 10.5 ms and 118.8 ms to reconstruct one image, respectively. 
These results evidenced that the deep beamformer was about three orders of magnitude faster than EISRCB in terms of FLOPs and faster than TEA on both CPU (5192 time) and GPU (11.3 time).

\begin{table*}
\centering
\caption{Energy spread area $A_{-3dB}$ and ISNR of the PAM images of two sources from the simulated and \emph{in vitro} tube experimental dataset for P4-1, L7-4 and CL15-7.}
\label{table:TwoSourcePerformance}
\setlength{\tabcolsep}{3pt}
\begin{tabular}{c|c|c|c|c|c|c|c}
\hline
\multirow{2}{*}{Data Type} & \multirow{2}{*}{Algorithm} & \multicolumn{3}{c|}{$A_{-3dB}$ ($mm^{2}$)} & \multicolumn{3}{c}{ISNR values (dB)} \\
\cline{3-8}
\multirow{2}{*}{} & \multirow{2}{*}{} 
& P4-1 & L7-4 & CL15-7 
& P4-1 & L7-4 & CL15-7\\
\hline
\multirow{3}{*}{\makecell[c]{simulated \\ data}} & deep beamformer 
& 8.6 $\pm$ 6.0 & 1.2 $\pm$ 0.7 & 0.7 $\pm$ 0.3
& 26.8 $\pm$ 4.4 & 21.1 $\pm$ 2.6 & 23.6 $\pm$ 2.3 \\
\multirow{5}{*}{} & EISRCB 
& 7.9 $\pm$ 5.7  & 1.0 $\pm$ 0.6 & 0.5 $\pm$ 0.3 
& 20.8 $\pm$ 3.6 & 22.5 $\pm$ 5.9 & 25.1 $\pm$ 5.1 \\
\multirow{5}{*}{} & TEA
& 10.6 $\pm$ 8.0 & 1.7 $\pm$ 0.8 & 1.0 $\pm$ 0.5
& 9.3 $\pm$ 0.7 & 11.8 $\pm$ 0.6 & 11.4 $\pm$ 0.6 \\
\hline
\multirow{3}{*}{\makecell[c]{experimental \\ data}} & deep beamformer 
& 12.2 $\pm$ 1.1 & 1.4 $\pm$ 0.3 & 2.9 $\pm$ 1.1
& 27.2 $\pm$ 1.2 & 31.9 $\pm$ 1.2 & 26.2 $\pm$ 1.3 \\
\multirow{5}{*}{} & EISRCB 
& 11.7 $\pm$ 1.3 & 1.1 $\pm$ 0.3 & 2.5 $\pm$ 1.2
& 24.1 $\pm$ 1.6 & 31.8 $\pm$ 2.1 & 25.6 $\pm$ 2.1 \\
\multirow{5}{*}{} & TEA
& 18.0 $\pm$ 0.5 & 3.8 $\pm$ 0.9 & 3.7 $\pm$ 1.6
& 9.4 $\pm$ 0.7 & 11.6 $\pm$ 0.9 & 10.6 $\pm$ 0.8 \\
\hline
\end{tabular}
\end{table*}

\subsection{Validation of the Deep Beamformer}
\subsubsection{Phantom experiments}
In the phantom experiments, the location of high-intensity region was also consistent in the reconstructed images using the deep beamformer and EISRCB (Fig.~\ref{fig:ExpResults}(a--f)). For single cavitation source, the images reconstructed by the deep beamformer presented no difference in the average $A_{-3dB}$ compared to the ones reconstructed by EISRCB (Table~\ref{table:SingleSourcePerformance}). Compared with TEA, the average $A_{-3dB}$ area and ISNR of the deep beamformer images was improved by 35.6\%--65.0\% and 17.1--19.0 dB, respectively, for all the transducers (Table~\ref{table:SingleSourcePerformance}). 
For multiple cavitation sources, the maximum difference in the average value of the $A_{-3dB}$ area and ISNR was less than 0.5 $mm^2$ and 3.1 dB. 
Compared with TEA, the average $A_{-3dB}$ area and ISNR of the deep beamformer images was improved by 21.6\%--63.1\% and 15.6--20.3 dB, respectively, for all the transducers (Table~\ref{table:TwoSourcePerformance}). 

The deep beamformer images showed small source localization deviation compared to the EISRCB images.
For single cavitation source, the proportion of the position deviation below 1 wavelength was 99.6\%, 93.6\%, and 95.6\% for P4-1, L7-4 and CL15-7, respectively, in the lateral direction, while the proportion of the position deviation below 4 wavelengths was 98.7\%, 99.3\%, and 98.8\% for P4-1, L7-4 and CL15-7, respectively, in the axial direction (Fig.~\ref{fig:ExpResults} (g--h)).
For multiple cavitation sources, the proportion of the lateral position deviation below 1 wavelength was 99.5\%, 95.1\%, and 96.6\% for P4-1, L7-4 and CL15-7 respectively, while the proportion of the axial position deviation below 4 wavelengths was 99.7\%, 82.8\%, and 80.3\% for P4-1, L7-4 and CL15-7, respectively (Fig.~\ref{fig:ExpResults} (i--j)). 

\begin{figure}[!t]
\centerline{\includegraphics[width=0.6\columnwidth]{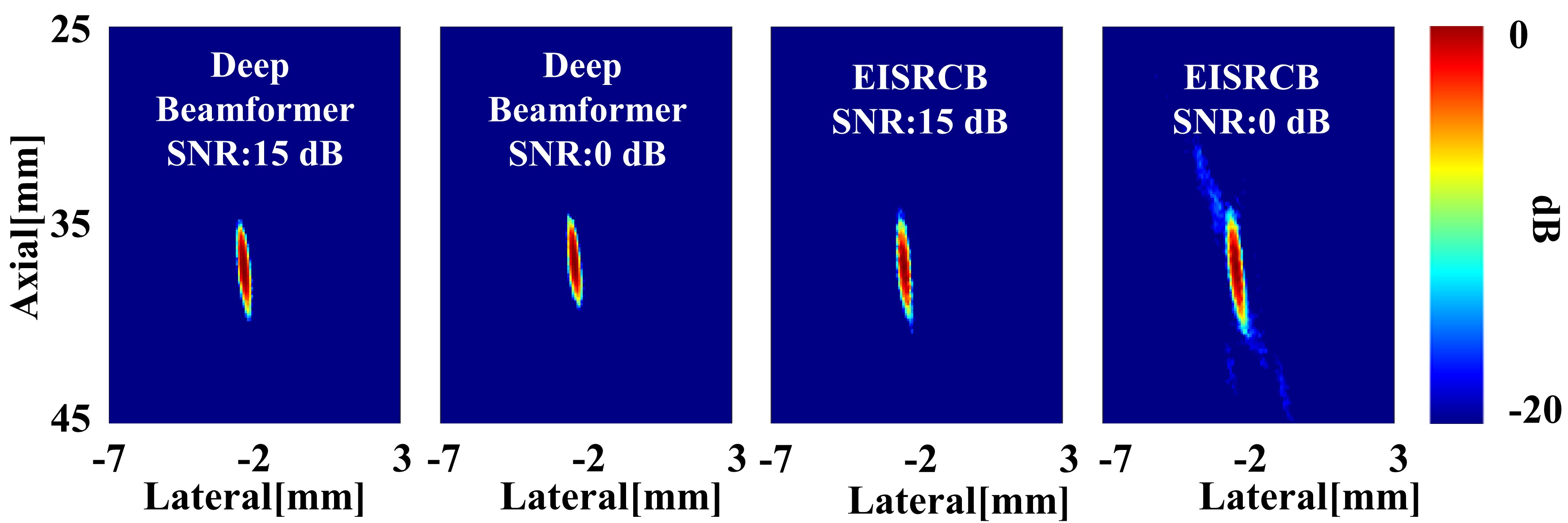}}
\caption{PAM images reconstructed by the deep beamformer and EISRCB using the simulated test set for L7-4 with noise of 15 dB noise and 0 dB.}
\label{fig:SimuNoiseTest} 
\end{figure}

\subsubsection{Mice experiments}
In the mice experiments, the deep beamformer also presented equivalent acoustic source localization capability as the EISRCB (Fig.~\ref{fig:ExpResults} (k)). The centroid of the cavitation source presented by the deep beamformer was at $\boldsymbol{r_c}=$ (2.3 mm, 24.6 mm), while it was at (2.4 mm, 24.8 mm) by EISRCB, indicating negligible source localization deviation between the two methods. The average ISNR and $A_{-3dB}$ of the deep beamformer images and the EISRCB images were similar (26.7 dB and 0.7 $mm^2$ vs 30.1 dB and 0.7 $mm^2$). Compared with TEA, the ISNR and $A_{-3dB}$ were improved by 19.6 dB and 58.8\% by the deep beamformer, respectively.
The deep beamformer achieved 10.5 ms image reconstruction speed on the GPU. Thus, real-time and high-quality localization of MB cavitation activities in the mouse tumors was possible (Supplementary Video 1).

\section{Discussion}
In this work, a deep beamformer for PAM has been presented. The deep beamformer was trained in a supervised learning approach with the EISRCB images used as the ground truth. This beamformer is switchable between different arrays (P4-1, L7-4, CL15-7) and directly maps the acoustic cavitation signals to PAM images with the same neural network.

We introduced both simulated and experimental (phantom and \emph{in vivo} mouse) data to train the neural network, which sums up to a total of 68400 acquisitions of the acoustic cavitation signals. For the simulation, we used the Marmottant and the Vokurka models to generate the acoustic emissions from MB clouds under stable and inertial cavitation, respectively. We further added the acoustic cavitation signals acquired in the experiments to include the signal characteristics originated from bubble-bubble interaction, which is currently difficult to simulate with a numerical model in a timely manner \cite{Coviello2015,Lu2018}. During data preparation, we considered various settings that are commonly used in MB-mediated FUS experiments \cite{gray2020}. The excitation frequency varied between 0.5--1.5 MHz, excitation duration varied between 58--98 $\mu$s, randomly distributed bubbles in the bubble clouds (single and multiple clouds) whose count varied between 20--100, and the center of the bubble cloud randomly fell in the full aperture range of the imaging arrays in the lateral direction and their respective nominal imaging depth (12.8--57.6 mm in the axial direction). We also varied the SoS of the medium between 1480--1600 m/s to help the beamformer adaptively reconstruct the images for various SoS values. In addition, we used the MBs from three different manufactures (Vevo, Sonovue, and Sonazoid) with varied acoustic pressures (PNP: 0.5--1.5 MPa) in the experiments. To improve the anti-noise capability of the network, the SNR of input RF signals was adjusted to 0--15 dB for each training epoch.
All these efforts were made to provide a comprehensive dataset for training the network.

\begin{figure*}[!t]
\centerline{\includegraphics[width=1.0\columnwidth]{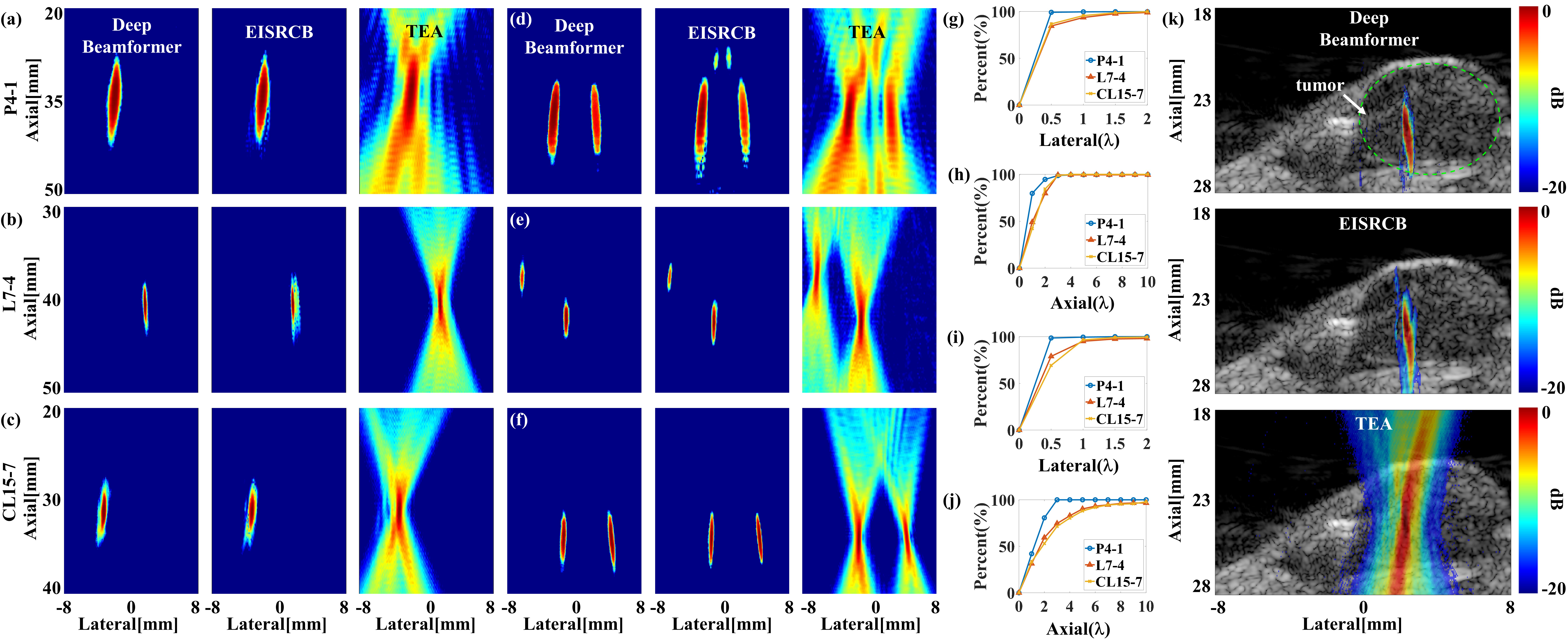}}
\caption{\emph{In vitro} experimental PAM images reconstructed by the deep beamformer, EISRCB, TEA using the test set for (a--c) phantoms and (d--f) double tubes with P4-1, L7-4 and CL15-7. Cumulative probability distribution of lateral position deviation and axial position deviation between the PAM images reconstructed by the deep beamformer and EISRCB for (g--h) single source and (i--j) multiple sources. The wavelength ($\lambda$) was evaluated at the center frequency of each transducer array. (k) \emph{In vivo} experimental PAM images reconstructed by the deep beamformer, EISRCB and TEA for CL15-7. The dynamic range of the B-mode images is 60 dB. The green dash circle outlines the tumor's boundary.}
\label{fig:ExpResults} 
\end{figure*}

While the other data-adaptive beamformers, i.e., DAX-RCB and RLPB, can produce high-quality PAM images as well, we found that EISRCB balances the best between the image quality, the sensitivity of the tuning parameters to the outcomes and the computational cost with our implementation. Compared to DAX-RCB, EISRCB had tight energy spread and less artifacts in the PAM images. Compared to RLPB, EISRCB was 9.3 times faster in execution time to reconstruct an image of 262144 pixels. In addition, EISRCB is relatively insensitive to the tuning parameters for consistent image reconstruction quality among various acquisitions. 
As an evidence, the same parameters were used in EISRCB for the \emph{in vitro} and \emph{in vivo} experiments, and both represented good artifacts suppression capability (Fig.~\ref{fig:ExpResults}). 
Therefore, we chose EISRCB to facilitate the production of the training dataset.

It turns out the network learned well the mapping between the cavitation signals and PAM images from the EISRCB method in our dataset. Hence, the images reconstructed by the deep beamformer were very similar to the ones by EISRCB. Their $A_{-3dB}$ area and the ISNR were very close and both showed reduced energy spread area and better artifacts suppression capability over TEA ((Table~\ref{table:SingleSourcePerformance} \& \ref{table:TwoSourcePerformance})). 
The investigated noise levels in the ultrasound signals (0-15 dB SNR) turned out to have a very small effect on the image reconstruction quality of the deep beamformer (Fig.~\ref{fig:SimuNoiseTest}), showing its capability of suppressing Gaussian noise in this SNR range.
In terms of source localization accuracy, more than 93.6\% of the position deviation was less than 1 wavelength in the lateral direction for all the arrays in both simulated and experimental data.
In the axial direction, more than 80.3\% of the deviation was less than 4 wavelengths.
Thus, the deep beamformer has comparable source localization accuracy as EISRCB.
We also tested the deep beamformer using the cavitation signals generated under the conditions not included in the training dataset (see Appendix). The results showed that the deep beamformer still can achieve similar image quality as EISRCB with slightly degraded source localization accuracy.

We built the deep beamformer based on GAN, which has been shown achieving better image quality than traditional convolutional neural network in ultrasound imaging \cite{tang2021plane,huang2021deep}.
To improve feature extraction from RF signals in the deep beamformer, two attention mechanisms were introduced: pixel attention for capturing global features and local aware attention for emphasizing high-frequency features. Traditional networks tend to overlook the relationship between channels due to the limited convolutional kernel size, and potentially neglecting important coherent acoustic emissions from MBs which is crucial for mapping cavitation activities in PAM. Additionally, networks may prioritize the learning of low-frequency features and ignore the high-frequency features which are important for differentiating the source location and the impact of SoS on the shape of the trace of RF signals. Pixel attention computes attention weights for each element of the feature maps, integrating channel and spatial information to focus on coherent acoustic emissions. Local aware attention amplifies values exceeding the average, effectively enhancing high-frequency feature representation. Without using these two attention mechanisms, the generator was found to wrongly place the cavitation sources in the images.

For the cavitation signals measured by different transducer models, a conventional generator cannot differentiate them. If raw RF signals were used as the input directly, the source location in PAM images reconstructed by the deep beamformer deviates remarkably from the true location. To solve this issue, mask vectors were used to introduce a prior information about transducer models to help the network better differentiate signal characteristics acquired by different arrays. 
This idea is similar to the concept of style transfer in Computer Vision~\cite{choi2018}.
By stacking mask vectors with the acoustic emission signals, a trained deep beamformer can be applied to process the data acquired by different transducer arrays. Thus, the flexibility of the deep beamformer was extended.

The computational complexity evaluated by FLOPs for PAM with the deep beamformer was reduced by more than 30 times and three orders of magnitude compared to TEA and EISRCB for the same image reconstruction task, respectively. Similar results are expected when compared to other data-adaptive algorithms, such as RLPB. 
During the implementation of the CUDA code for EISRCB, we noticed two limitations preventing it achieving faster image reconstruction speed on the GPU compared to the CPU. The limitations are: 1) cuSolver, the library used for singular value decomposition (SVD) which is required, could not execute parallel operations for matrices larger than 64 by 64, resulting in slower SVD execution on GPU compared to CPU. 2) parallelizing EISRCB requires significant amount of memory (about 256 GB) and dividing the image to sub-image showed negligible improvements in image reconstruction speed on the GPU.
Therefore, we were not able to directly compare the required time to reconstruct one image by the deep beamformer and EISRCB on the GPU.
However, such a direct comparison may not be necessary as the deep beamformer was already 11.3 times faster than TEA and achieved better image quality than the latter.
Thus, we showed that in the treatment of the mice tumors  (Fig.~\ref{fig:ExpResults} (k)), the deep beamformer allows seamless combination of B-mode and PAM images for real-time and accurate localization of MB cavitation activities (Supplementary Video 1). Taken together, we showed that the deep beamformer has the potential of being an alternative to data-adaptive beamformers for real-time and high-quality PAM, which may contribute to more accurate localization of cavitation sources, and consequently improve therapeutic effects and reduced risks of damaging normal tissues in cavitation-based FUS therapy~\cite{Coviello2015}.

There are some limitations in this work. Firstly, the acoustic signals simulated by the Vokurka and Marmottant model may not fully capture the characteristics of the acoustic emissions of a cavitating bubble cloud, because bubble-bubble interaction was not considered in both models. While we have used experimental data to complement these missing signal characteristics, it remains to examine whether a more realistic model incorporating bubble-bubble interaction is necessary to further improve the performance of the deep beamformer. Secondly, we only considered PAM with planar arrays. Thus, the deep beamformer may not be directly applicable for arrays of other shapes for PAM, such as convex arrays.
Nonetheless, the neural network may be extended to adapt to various array geometries by using the style transferring approach which has been used in this work for representing different planar arrays (P4-1, L7-4, CL15-7), with carefully prepared dataset for training the network. 

\section{Conclusion}
In this work, we developed a deep beamformer that can reconstruct PAM images of equivalent quality as the ones produced by the EISRCB algorithm with much less computational cost, for localizing one cavitation source and multiple cavitation sources. Results also showed that by introducing mask vectors, high-quality PAM images can be reconstructed directly from the RF signals for different transducer arrays using the same neural network. Using deep learning based beamformers for PAM is promising for real-time monitoring and accurate localization of MB cavitation activities.

\section*{CRediT authorship contribution statement}
\textbf{Yi Zeng:} Writing – original draft, Methodology, Investigation, Software, Validation, Data curation, Visualization, Formal analysis. 
\textbf{Jinwei Li:}  Methodology, Resources. 
\textbf{Hui Zhu:} Methodology, Data curation.  
\textbf{Shukuan Lu:} Writing – review \& editing, Methodology, Investigation. 
\textbf{Jianfeng Li:} Writing – review \& editing, Methodology, Investigation.
\textbf{Xiran Cai:} Writing – review \& editing, Conceptualization, Supervision, Project administration, Funding acquisition.

\section*{Declaration of competing interest}
The authors declare that they have no known competing financial interests or personal relationships that could have appeared to influence the work reported in this paper.

\section*{Data availability}
Data will be made available on request.

\section*{Acknowledgments}
This work was supported by the Natural Science Foundation of Shanghai (24ZR1450700), National Natural Science Foundation of China (12404533).

\section*{Appendix}
 \label{sec:Appendix}

 \begin{figure}[!t]
 \centerline{\includegraphics[width=0.55\columnwidth]{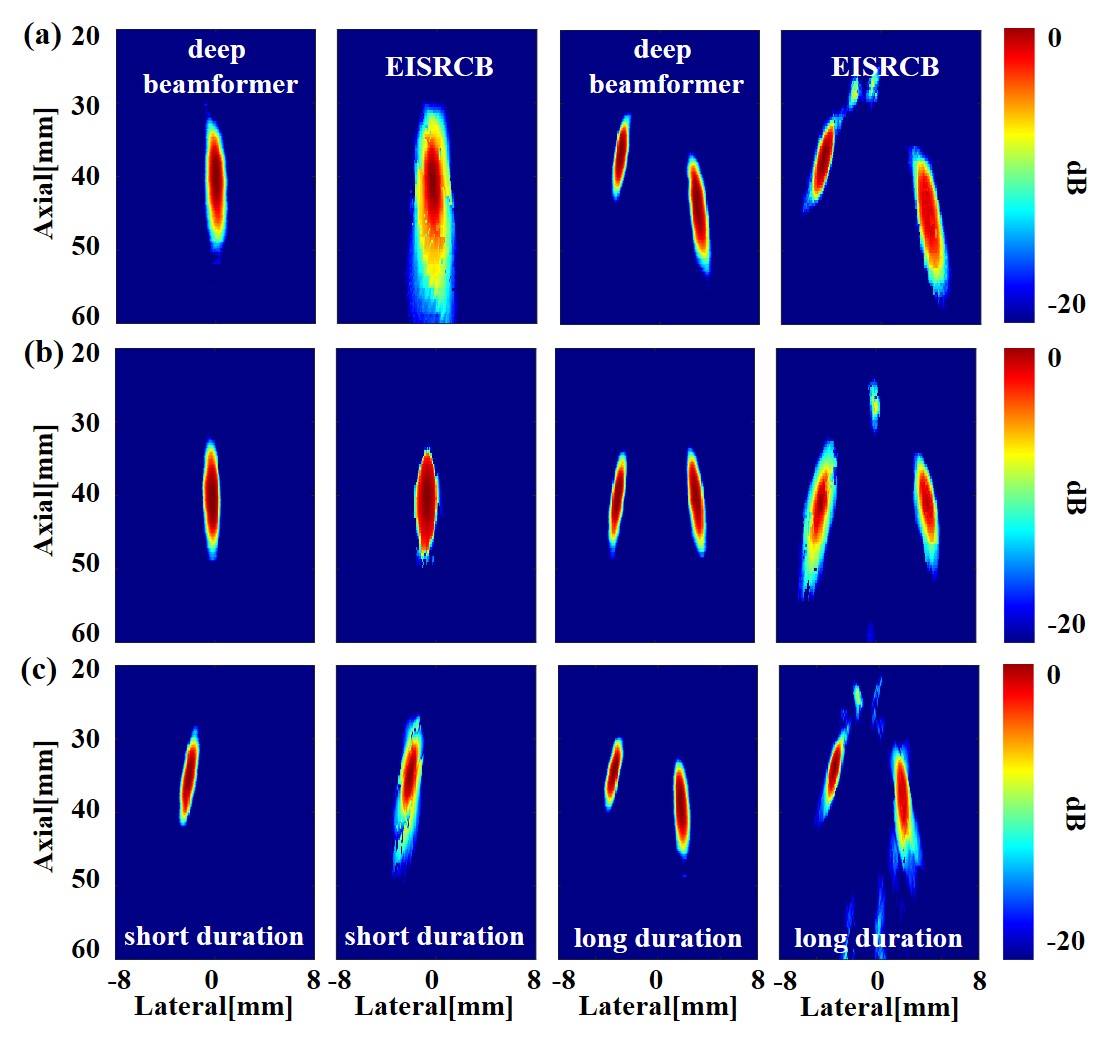}}
\caption{PAM images reconstructed with the simulated cavitation signals for P4-1 under the conditions not included in the training dateset, including (a) short pulses (10 cycles) excitation, (b) 2 MHz excitation frequency, (c) long (154 $\mu$s, 3072 samples) and short (51 $\mu$s, 1024 samples) acquisition duration.}
\label{fig:DifferentConditions} 
 \end{figure}

 We also evaluated the performance of the deep beamformer for the cavitation signals, received by the P4-1, generated under the conditions not included in the training dataset.
 These conditions include using short excitation pulses (10 cycles) to generate the cavitation signals, higher excitation frequency (2 MHz), and long  (3072 samples) and short (1024 samples) duration. Evaluation metrics were calculated using 100 simulated data for each condition.

1) Short pulses (Fig.~\ref{fig:DifferentConditions}(a)): 
The difference of the averaged ISNR and $A_{-3dB}$ between the deep beamformer and EISRCB was 4.9 dB (5.4 dB) and 0.9 $mm^2$ (0.9 $mm^2$) for single (double) source(s), respectively.
The percent of source position deviation below 1 wavelength was 98.0 \% in the lateral direction, while the proportion of the deviation below 4 wavelengths was 90.5 \% in the axial direction.

2) 2 MHz excitation frequency (Fig.~\ref{fig:DifferentConditions}(b)): The difference of the averaged ISNR and $A_{-3dB}$ between the deep beamformer and EISRCB was 3.4 dB (2.0 dB) and 0.9 $mm^2$ (2.1 $mm^2$) for single (bubble) source(s), respectively. The percent of source position deviation below 1 wavelength was 95.0\% in the lateral direction, while the proportion of the deviation below 4 wavelengths was 74.5\% in the axial direction.

3) Long and short acquisition duration (Fig.~\ref{fig:DifferentConditions}(c)): For long (3072 samples) and short duration (1024 samples), the signals were trimmed down and zero-padded to 2048 samples to fit to the input dimension of the deep beamformer. The difference of the averaged ISNR and $A_{-3dB}$ between the deep beamformer and EISRCB was 2.2 dB (2.7 dB) and 0.3 $mm^2$ (0.2 $mm^2$) for single (double) bubble source(s), respectively. The percent of source position deviation below 1 wavelength was 99.6 \% in the lateral direction, while in the axial direction, it was 99.0 \% below 4 wavelengths.

  \bibliographystyle{elsarticle-num-names} 
  \bibliography{Ultrasonics/bibfiles}



\end{document}